\documentclass[11pt]{article}
\usepackage{coling2018}
\usepackage{times}
\usepackage{url}
\usepackage{latexsym}
\usepackage{amsmath}
\usepackage{graphicx}
\usepackage[justification=centering]{caption}
\renewcommand\vec[1]{\overrightarrow{#1}}
\newcommand\cev[1]{\overleftarrow{#1}}
\usepackage{xcolor}
\usepackage{soul}
\usepackage{natbib}
\usepackage{dsfont}
\usepackage{multirow}
\usepackage{subcaption}

\title{KDSL: a Knowledge-Driven Supervised Learning Framework for \\Word Sense Disambiguation}
\author{Shi Yin\textsuperscript{1},  Yi Zhou\textsuperscript{2}, Chenguang Li\textsuperscript{1}, Shangfei Wang\textsuperscript{1}, Jianmin Ji\textsuperscript{1}, Xiaoping Chen\textsuperscript{1}, Ruili Wang\textsuperscript{3}\\
\textsuperscript{1}School of Computer Science and Technology, University of Science and Technology of China, China\\
\textsuperscript{2}School of Computing, Engineering and Mathematics, University of Western Sydney, Australia\\
\textsuperscript{3}Institute of Natural and Mathematical Sciences,  Massey University (Albany Campus), New Zealand\\
davidyin@mail.ustc.edu.cn, yizhoujoey@gmail.com, cg0808@mail.ustc.edu.cn\\
\{sfwang,jianmin,xpchen\}@ustc.edu.cn, Ruili.WANG@MASSEY.AC.NZ
}
\begin{document}

\maketitle

\begin{abstract}
We propose KDSL, a new word sense disambiguation (WSD) framework that utilizes knowledge to automatically generate sense-labeled data for supervised learning. First, from WordNet, we automatically construct a semantic knowledge base called DisDict, which provides refined feature words that highlight the differences among word senses, i.e., synsets. Second, we automatically generate new sense-labeled data by DisDict from unlabeled corpora. Third, these generated data, together with manually labeled data and unlabeled data, are fed to a neural framework conducting supervised and unsupervised learning jointly to model the semantic relations among synsets, feature words and their contexts. The experimental results show that KDSL outperforms several representative state-of-the-art  methods on various major benchmarks. Interestingly, it performs relatively well even when manually labeled data is unavailable,  thus provides a potential solution for similar tasks in a lack of manual annotations.
\end{abstract}

\section{Introduction}

Word sense disambiguation (WSD) is the task to identify the sense of
a word under certain context. It is one of the central tasks for
understanding natural languages. WSD has been widely used in many
basic natural language processing (NLP) tasks or downstream applications,
such as sentiment analysis
\citep{huang2012improving} and machine translation \citep{neale2016word}.

Approaches for WSD are divided into two groups, i.e., (semi)
supervised learning
\citep{lee-ng:2002:EMNLP02,Zhi2010It,DBLP:journals/corr/KagebackS16,Iacobacci2016Embeddings,DBLP:conf/coling/YuanRDEA16,DBLP:conf/conll/MelamudGD16,Raganato2017Neural}
and knowledge-based approaches
\citep{Lesk1986Automatic,Banerjee2003Extended,Agirre2009Personalizing,Miller2012Using,Moro2014Entity,Basile2014An}.
In general, the former approaches perform better than the latter in
most benchmarks. However, most supervised learning approaches for WSD are heavily dependent on the amount of sense-labeled data. Unfortunately, sense-labeled data is far from adequate for supervised systems to perform well due to the high cost of manual annotations. For synsets never occurred in the training corpora, these methods can not learn to make plausible predictions.

Motivated by this, we propose KDSL, a new framework to combine
supervised learning and knowledge-based approaches for WSD
 by automatically generating sense-labeled data from
explicit knowledge bases as the training dataset for supervised
learning. More precisely, we first build a high quality semantic
knowledge base from WordNet \citep{DBLP:journals/cacm/Miller95} that
highlights the differences among word senses. Then, we utilize this
knowledge base to generate sense-labeled data from raw sentences.
Finally, these automatically generated data are fed to a neural
network to model the semantic relationships among word senses,
feature words and their contexts.

\begin{figure*}
\begin{minipage}[t]{0.5\linewidth}
\centering
\includegraphics[width = 60 mm, height = 58 mm]{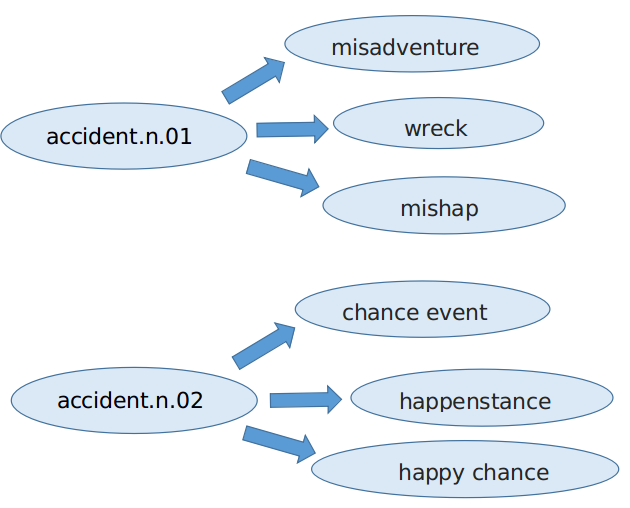}
\subcaption{Two synsets of the word ``accident" in DisDict}
\label{english}
\end{minipage}%
\begin{minipage}[t]{0.5\linewidth}
\centering
\includegraphics[width = 60 mm, height = 58 mm]{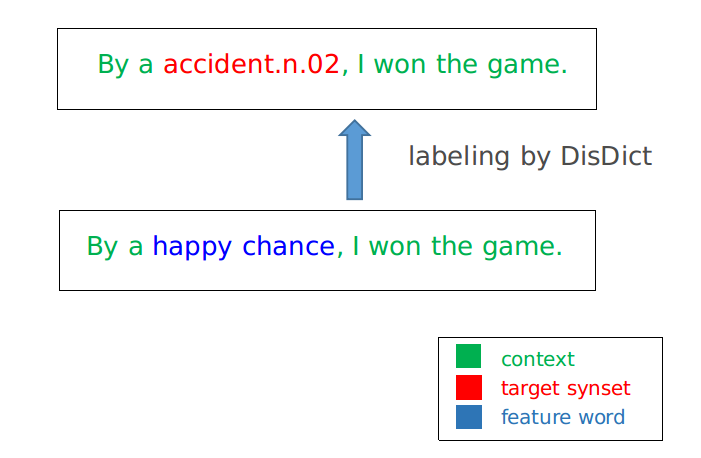}
\subcaption{Since ``happy chance" serves as a feature word for ``accident.n.02", labeling the context of ``happy chance" \\with ``accident.n.02" as a new supervised learning instance}
\label{data_gen}
\end{minipage}
 \caption{Generating labeled data for ``accident.n.02" by DisDict from plain sentences}
\end{figure*}

For the first step, we construct DisDict, a semantic KB customized
for WSD, which is automatically extracted from WordNet by a
statistic model. It selects simple feature words to highlight the
differences among word senses, i.e., synsets. DisDict contains a
number of triples of the form (synset, feature words, confidence
score) for all synsets in WordNet 3.0, which contains a total number
of 117659 synsets covering nouns, verbs, adjectives and adverbs. The
feature words are selected based on two criteria. Firstly, they
should have similar semantics with the synset. Secondly, different
from previous semantic KBs such as WordNet and ConceptNet, DisDict
specifically aims at WSD, i.e., to highlight the differences among
different synsets during knowledge extraction. For instance,
depicted in Figure \ref{english}, the word ``accident" has two
synsets, namely ``accident.n.01" for ``an unfortunate mishap;
especially one causing damage or injury", and ``accident.n.02"  for
``anything that happens suddenly or by chance without an apparent
cause".  For the synset ``accident.n.01", DisDict chooses
``misadventure",  ``wreck", ``mishap" as the feature words with
highest confidence scores, while for ``accident.n.02", the top three
feature words are ``chance event", ``happenstance" and ``happy
chance". Clearly, those two sets of feature words provide
significant discriminative information  between these two synsets.

The second step is to generate sense-labeled data automatically by
DisDict from raw sentences. Since a synset is semantically similar
to its feature words in DisDict, if one of these words occurs in a
sentence, we label the context with the synset as a new instance.
For instance, depicted in Figure \ref{data_gen}, since ``happy
chance" serves as a feature word for ``accident.n.02", the context
in which ``happy chance" occurs is labeled with ``accident.n.02" and
can be fed into supervised learning. In this way, we can generate
much new labeled data for target synsets.

\begin{figure*}
\begin{minipage}[t]{0.5\linewidth}
\centering
\includegraphics[width = 80 mm, height = 62 mm]{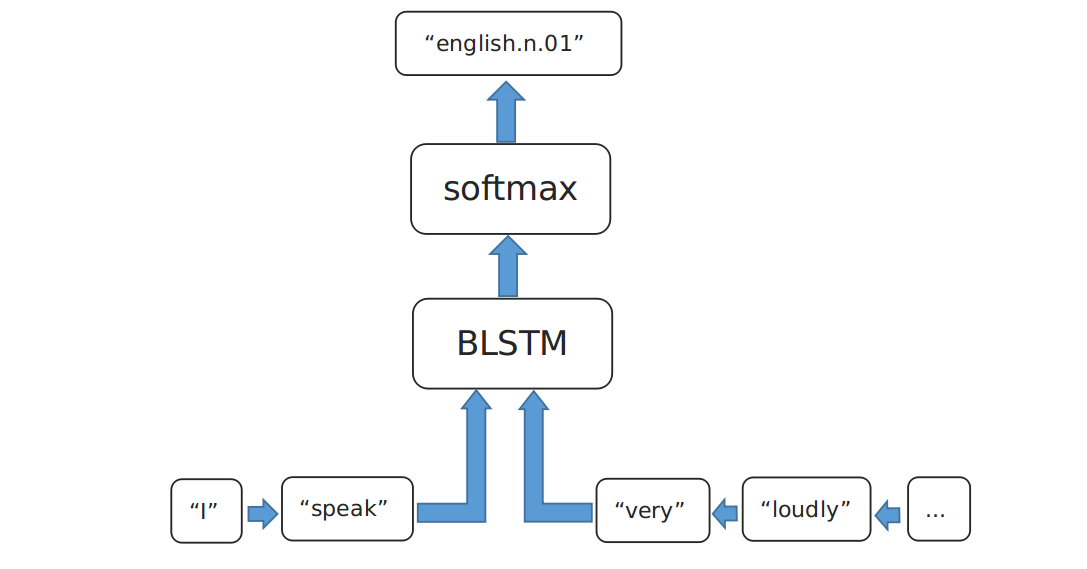}
\subcaption{Supervised learning to predict synset ``english.n.01" by the left context ``I speak" and right context ``very loudly ..." of ``English" in sense-labeled corpora }\label{fig1}
\end{minipage}%
\begin{minipage}[t]{0.5\linewidth}
\centering
\includegraphics[width = 80 mm, height = 62 mm]{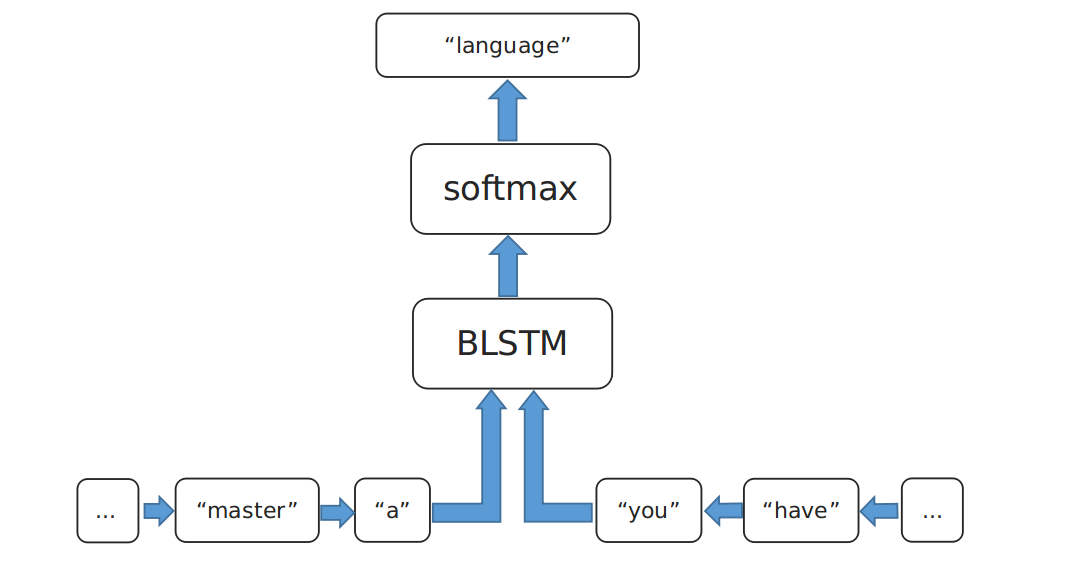}
\subcaption{Unsupervised learning to predict word ``language" by its left context ``To master a" and right context ``you have to practice listening and speaking ..." in unlabeled corpora}\label{fig2}
\end{minipage}
 \caption{Supervised and unsupervised learning for WSD}
\end{figure*}

The final step is to design a neural framework conducting learning on these generated data, together with manually labeled data and unlabeled data. Depicted in Figure \ref{fig1}, given a sentence and a word in
it to be disambiguated, the neural network takes the left context
before the word and the right context after the word as the input,
and uses a binary long short-term memory (BLSTM) encoder to encode
them as a fixed-length context embedding, which is fed into a fully
connected network with multi softmax outputs for predictions. As a
supervised learning task, this encoder is trained jointly on data automatically generated by DisDict as well as data manually labeled to predict the
proper synsets by their contexts. We set a param to
control the ratio of samples from the two data sources. To
improve the generalization ability, we also design an unsupervised
learning task, i.e., training the encoder on unlabeled corpora to
predict words by their left and right contexts, as depicted in
Figure \ref{fig2}.

We conduct empirical evaluations on various major WSD datasets and
our method outperforms a number of representative approaches.
Experiments show that incorporating supervised learning on  the data
generated by DisDict improves the performance for WSD.  Even when
there is no sense-labeled data, our work also performs well and
beats MFS, which is a state-of-the-art knowledge-based WSD method. Our approach illustrates that the combination of semantic knowledge and unlabeled data is useful to generate high quality sense-labeled data and provides a potential solution for similar tasks without manually labeled data.

\section{Related Work}

In this section, we will briefly review previous approaches about supervised WSD, knowledge-based WSD, combined methods and data generation strategies for this task.
\subsection{Supervised WSD}
Supervised WSD is trained on sense-labeled corpora. The labels and features for training are extracted either manually or automatically. \citet{Zhi2010It} utilized surrounding words, POS tags of surrounding words and local collocations as features and trained a classifier for WSD. \citet{DBLP:conf/acl/RotheS15} leveraged WordNet to generate synset embeddings from word embeddings and convert them into features of a supervised learning system. \citet{DBLP:journals/corr/KagebackS16} proposed an approach based on bidirectional LSTM to model sequence of words surrounding the target word without hand-crafted features. \citet{Iacobacci2016Embeddings} published a full evaluation study on equipping supervised WSD with word embeddings. To alleviate the lack of sufficient manually labeled corpora, \citet{DBLP:conf/coling/YuanRDEA16} proposed a semi-supervised framework with label propagation to expand training corpora. \citet{DBLP:conf/conll/MelamudGD16} proposed a generic model for representation of context, i.e., context2vec, and fed it into a classifier for WSD.  \citet{USLU18.736} proposes fastsense, a neural WSD model with high learning efficiency.
\subsection{Knowledge-based WSD}
Knowledge-based approaches  rely on manually constructed human knowledge base. \citet{Lesk1986Automatic} proposed definition (gloss) overlap measure, i.e., to calculate overlaps among the definitions of the target word and those surrounding it in the given context to determine word sense. It was enhanced by \citet{Banerjee2003Extended} to take definitions of related words into consideration. \citet{Chen2011CombiningCA} combined both WordNet and ConceptNet to judge word sense. Taking advantages of distributional similarity \citep{Miller2012Using,DBLP:conf/coling/BasileCS14,DBLP:conf/emnlp/ChenLS14,DBLP:conf/lrec/Camacho-Collados16} has also been shown effective. \citet{Agirre2009Personalizing,DBLP:conf/acl/GuoD10,DBLP:journals/coling/AgirreLS14,Moro2014Entity,DBLP:conf/acl/WeissenbornHXU15,DBLP:journals/coling/TripodiP17} modeled knowledge bases as graphs, i.e., words as nodes and relations as edges. The senses preferences of each word are updated iteratively according to certain graph-based algorithms. \citet{pasini2018two} proposed two knowledge-based methods for learning the distribution of senses.

\subsection{Combined methods and data generation strategies}
\citet{DBLP:conf/acl/RotheS15} leveraged WordNet to generate synset embeddings from word embeddings and convert them into features of a supervised learning system; \citet{Raganato2017Neural} introduced several advanced neural sequence learning models to WSD and design a multi-task mechanism to predict synsets as well as their coarse-grained semantic labels. \citet{DBLP:conf/conll/TaghipourN15} proposed OMSTI, a sense-labeled corpus generated through the disambiguation of a multilingual parallel corpus; \citet{pasini-navigli:2017:EMNLP2017} proposed Train-O-Matic, a data generation strategy based on random walk in WordNet.
\section{Models}
\subsection{Problem Formalization}
Suppose there is a sentence $c$ with words in order: $w_1$, $w_2$, ..., $w_L$, each of which is tagged with its POS. For instance, in a sentence ``Knowledge is power", $w_1$=``knowledge\_NOUN", where the suffix ``\_NOUN" means its POS is noun. For each $w_i$ in $c$, there is a set of candidate synsets $CS(w_i)=\{s_1(w_i),s_2(w_i),  ...,s_K(w_i)\}$. If two synsets are both candidate synsets for a certain word, it is called that they have competitive relations in this paper. The goal of word sense disambiguation (WSD) is to identify the correct synset of $w_i$ given the context $c$. For a corpus $T$ with sentences $c_1$, $c_2$, ..., $c_N$, the collection of all target words to be disambiguated in $T$ is denoted as $TW$, the collection of all candidate synsets for words in $TW$ is denoted as $TS$. And in DisDict, there are several feature words to interpret the semantics of each synset, the collection of all feature words is denoted as $FW=\cup_i FW(s_i)$, while $FW(s_i)$ is the collection of feature words of synset $s_i$.

\subsection{Knowledge Base for WSD: DisDict}
\subsubsection*{Motivations}
For our work, we need a semantic KB to generate sense-labeled data. However, existing semantic KBs are hard to use directly. There are two main disadvantages:

a) Coarse-grained. Some semantic KBs, e.g., ConceptNet, only provide word (or phrase) level knowledge and do not distinguish different potential senses of a given word (phrase) explicitly;

b) (Partially) unstructured. WordNet and BabelNet \citep{DBLP:journals/ai/NavigliP12} provide glosses of synsets by unstructured texts which are hard to encode and utilize by neural models.

Motivated by these disadvantages, we propose DisDict, a semantic KB aiming at WSD. The ideas for DisDict are also also twofold:

a) Establishing synset level semantic knowledge;

b) Extracting high-quality semantic information from (partially) unstructured knowledge to highlight the distinction among candidate synsets. In DisDict, only words having high statistic correlations with the target synset are selected as its feature words. These words are tagged with confidence scores by a statistic model.  Noisy words or words with little discriminative information for synsets are removed.
\begin{table*}
    \begin{center}
    \resizebox{\textwidth}{50mm}{
        \begin{tabular}{|c|c|c|} \hline
            (player.n.01, playmaker, 0.11)& (arrive.v.01, flood in, 0.21) & \\
            (player.n.01, seeded player, 0.11)& (arrive.v.01, plump in, 0.21) &(brainy.s.01, brainy, 0.52) \\
            (player.n.01, dart player, 0.11)& (arrive.v.01, drive in, 0.16) & \\
            (player.n.01, most valuable player, 0.11)& (arrive.v.01, move in, 0.16) &(brainy.s.01, smart as a whip, 0.36) \\
            (player.n.01, volleyball player, 0.11)& (arrive.v.01, get in, 0.05) & \\
            (player.n.01, pool player, 0.09)& (arrive.v.01, come in, 0.04) & (brainy.s.01, impressive, 0.06)\\
            (player.n.01, lacrosse player, 0.09)&  (arrive.v.01, draw in, 0.04)& \\
            (player.n.01, grandmaster, 0.09)&(arrive.v.01, set down, 0.04)  & (brainy.s.01, unusual, 0.05)\\
            (player.n.01, scorer, 0.09)& (arrive.v.01, roll up, 0.04) & \\
            (player.n.01, billiard player, 0.09)& (arrive.v.01, attain, 0.03) & \\ \hline
            (musician.n.01, flutist, 0.1)&  &\\
            (musician.n.01, vibist, 0.1)& (arrive.v.02, succeed, 0.41)& \\
            (musician.n.01, accompanist, 0.1)&  &(brilliant.s.01, transcendent, 0.57) \\
            (musician.n.01, harmonizer, 0.1)&  & \\
            (musician.n.01, gambist, 0.1)& (arrive.v.02, come through, 0.4) & \\
            (musician.n.01, carillonneur, 0.1)&  & (brilliant.s.01, surpassing,0.43)\\
            (musician.n.01, recorder player, 0.1)&  & \\
            (musician.n.01, harper, 0.1)& (arrive.v.02, win, 0.2) & \\
            (musician.n.01, keyboardist, 0.1)&  & \\
            (musician.n.01, accompanyist, 0.1)&  & \\ \hline
        \end{tabular}}
        \caption{A glance of DisDict}
        \label{disdictnf}
    \end{center}
\end{table*}

\subsubsection*{Construction of DisDict}

To build DisDict, firstly, words from Synonymy, Hypernymy/Hyponymy and Gloss in WordNet are harvested as potential feature words. A word in a synset's gloss is likely to be its feature word only if it has the same POS with the synset. For instance, since the gloss of synset ``english.n.01" is ``the people of England", then ``people" and ``England" are considered as potential feature words of ``english.n.01".

Then a statistic model is implemented to select words having high statistical correlations with the synsets from potential feature words. This model is similar to PMI (point-wise mutual information) and was firstly proposed by \citet{Wettler1993Computation} to compute word associations.  For each (synset, feature word) pair ($s_i$, $w_j$), $r_{ij}$ represents the strength of their correlations:
\begin{small}
\begin{equation}
r_{ij}=\frac{p(s_i,w_j)}{p(s_i)p(w_j)^\tau}=\frac{p(s_i,w_j)}{\sum_w{p(s_i,w)}\sum_s{p(s,w_j)}^\tau}
\label{pmi}
\end{equation}
\end{small}

where $p(s_i,w_j)$ is the co-occurrence probability, which is proportional to the times $s_i$ associated with $w_j$ in WordNet. For instance, when traversing WordNet, if $w_j$ occurs twice in the gloss of $s_i$, then $p(s_i, w_j)\gets p(s_i, w_j)+2$. $\tau$ is an adjustable parameter to control the effect of word frequencies. For each target synset $s_i$, only a small number ($<=N_f$) of feature words with the highest $r_{ij}$ are preserved, others are dropped out. Confidence scores for these $N_f$ feature words are normalized such that their sum is 1.0， which is:
\begin{small}
\begin{equation}
\label{freq_fw}
\begin{split}
r_{ij}\gets \frac{r_{ij}}{\sum_{w_{j'}\in FW(s_i)}r_{ij'}},w_j \in FW(s_i)
\end{split}
\end{equation}
\end{small}
DisDict is organized as a number of triples, i.e., $(s_i, w_j, r_{ij})$. Table \ref{disdictnf} provides a glance of DisDict, i.e., three competitive synsets pairs in DisDict ($N_f=10$), i.e., ``player.n.01" (`` a person who participates in or is skilled at some game") and ``musician.n.01" (``someone who plays a musical instrument (as a profession)"); ``arrive.v.01" (``make a prediction about; tell in advance") and ``arrive.v.02" (``succeed in a big way; get to the top"); ``brainy.s.01" (``having or marked by unusual and impressive intelligence") and ``brilliant.s.02" (``of surpassing excellence"). Synsets in the same column have competitive relations with each other.
\begin{tiny}

\end{tiny}
\subsubsection*{Sense-Labeled Data Generation by DisDict}

The second step is to generate sense-labeled data automatically by DisDict from raw sentences. Since in DisDict a synset is semantically similar to its feature words, if one of them occurs in a sentence, label the sentence with the synset as a new instance. Depicted in Figure \ref{data_gen}, since ``happy chance" serves as a feature word for ``accident.n.02", the context in which ``happy chance" occurs is labeled with ``accident.n.02" and can be fed into supervised learning.  In this way, we can generate much new labeled data for target synsets.

Synsets in corpora follow a certain frequency distribution. Different synsets may have different frequencies. Since we do not know this distribution a priori, we can only design a model to simulate it, i.e., the frequencies for synset $s_i$ are approximately calculated as:
\begin{small}
\begin{equation}
\label{freq}
\begin{split}
f(s_i)\propto\sum_{w_j, s_i\in CS(w_j)}f(w_j) \cdot p^{l_{ij}}
\end{split}
\end{equation}
\end{small}
where $w_j$ is a word of which $s_i$ is one of the candidate synsets, i.e., $s_i\in CS(w_j)$; $l_{ij}$ denotes the ranking of $s_i$  among synsets in $CS(w_j)$, e.g., the ranking of ``english.n.01" is 1 among the candidate synsets of "English\_Noun" in WordNet; $f(w_j)$ is the frequency of $w_j$ which is counted in a large corpus; $p$ is an adjustable param to control the bias towards synsets with high rankings.

The instances for $s_i$ are generated by its feature words, i.e.,  $FW(s_i)$. The number of instances contributed by $w_k \in FW(s_i)$ for $s_i$, denoted as  $f(s_i, w_k)$, is set as:

\begin{small}
\begin{equation}
\label{freq_fw}
\begin{split}
f(s_i, w_k)= f(s_i) \cdot r_{ik}, w_k \in FW(s_i)
\end{split}
\end{equation}
\end{small}

\subsection{Learning Framework}
The learning process has two parts:

a) Supervised Learning: training a model to predict synsets by contexts in sense-labeled corpora. As shown in Figure \ref{fig1}, suppose there is a sentence ``I speak English very loudly..." where the word ``English" is to be disambiguated. The model is trained to make prediction of ``english.n.01" over all synsets given the left (``I speak") and right (``very loudly ...") context of ``English".

b) Unsupervised learning: training the model in a) to predict words by contexts in large unlabeled corpora. As shown in Figure \ref{fig2}, suppose there is a sentence ``To master a language, you have to practice listening and speaking ..." where the word ``language" is one of the feature words in DisDict, the model is trained to make prediction of ``language" given its left (``To master a") and right (``you have to practice listening and speaking ....") context. This training process promotes the ability to model context and extract semantic features, which improves the generalization performance.

During training, instances are sampled from manually labeled data and data generated by DisDict.  We set a param to
control the ratio of samples from the two data sources.

\begin{table*}[t]
    \centering
    \begin{tabular}{|c ||c | c | c | c ||c |c |c |c |c |} \hline
                  \multirow{2}*{Methods}&  \multicolumn{4}{|c||}{Test Datasets}  & \multicolumn{5}{|c|}{Concatenation of All Test Sets}  \\ \cline{2-10}
            & SE2 &  SE3 & SE13   & SE15 & Nouns & Verbs &  Adj. & Adv.  & All\\ \hline
            IMS & 70.9 & 69.3 & 65.3 & 69.5   & 70.5 & 55.8 &  75.6 & 82.9 & 68.9 \\
            IMS{-s}+emb& \textbf{72.2} & 70.4 & 65.9 & 71.5   & 71.9 & 56.6 &  75.9 & \textbf{84.7} & 70.1 \\
            Context2vec& 71.8 & 69.1 & 65.6 & 71.9   & 71.2 & \textbf{57.4} &  75.2 & 82.7 & 69.6 \\
            Le et al. (2017)& 70.0 & - & 66.6 & -   & - & - &  - & - & - \\
            Raganato et al. (2017)& 72.0 & 69.4 & 66.4 & 70.8   & 71.6 & 57.1 &  75.6 & 83.2 & 69.9 \\ \hline
            \multicolumn{10}{|c|}{Ours}\\ \hline
            MLab& 69.2& 68.3& 66.1& 67.4& 69.4& 54.6& 75.7& 82.4& 68.0\\
            MLab+ULab & 70.2& 69.9& 69.3& 72.9 & 72.3& 56.8& 77.4& 81.1& 70.4\\
            MLab+ULab+MFS & 70.8& 69.9& 69.8& \textbf{73.0} & 72.8& 56.8& 77.3& 81.2& 70.7\\
            MLab+DisDict+ULab& 72.0   & 70.5  &  \textbf{70.9}  & 72.7   & 74.3  & 55.6   & 77.7    &  82.4  &  71.4  \\
            MLab+DisDict+ULab+MFS& 72.0 & \textbf{71.2} & \textbf{70.9} & 72.9   & \textbf{74.4} & 56.0 &  \textbf{78.3} & 82.1 & \textbf{71.7} \\
             \hline
    \end{tabular}
    \caption{F1-scores (\%) for English all-words fine-grained WSD with the utilization of manually labeled training data}
    \label{result}
\end{table*}
\subsection{Neural Model}

LSTM \citep{Hochreiter1997Long,DBLP:journals/nn/GravesS05} is a gated type of recurrent neural network (RNN), which is a powerful model for NLP.  We follow to choose BLSTM  \citep{DBLP:journals/nn/GravesS05} as our basic neural encoder. As shown in Figure 1, suppose in a sentence $c$ with words $w_1,w_2,...,w_t,w_{t+1},...,w_{N}$, $w_t$ is the target word to be disambiguated. We arrange words surrounding $w_t$ in order, which are denoted as $left\_word=[w_{t-T},w_{t-T+1},...,w_{t-1}]$ and $right\_word=[w_{t+1},w_{t+2},...,w_{t+T}]$, where $T$ is the maximal distance we consider. Then the two groups of word sequences are fed into a BLSTM structure with two LSTMs in different directions. The words on the left side of the target word are fed into a left-to-right LSTM while those on the right side of target are fed into right-to-left LSTM. The left-to-right LSTM generates a sequence of hidden state vectors $[\vec{h_1},...,\vec{h_T}]$ and the right-to-left LSTM generates a sequence of hidden state vectors $[\cev{h_1},...,\cev{h_T}]$. Then we get two feature vectors $\vec{h_l}$ and $\cev{h_r}$ for left and right context of $w_t$:
\begin{small}
\begin{equation}
\vec{h_l}=\vec{h_T},\cev{h_r}=\cev{h_T}
\end{equation}
\end{small}
Since the left and right contexts of $w_t$ do not contain $w_t$ itself, they are denoted as $c \backslash \{w_t\}$. The vector representation for context $c \backslash \{w_t\}$, i.e., $\vec{c_{\backslash t}}$, is calculated by a fully connected module which takes the concatenation of $\vec{h_l}$ and $\cev{h_r}$ as input:
\begin{small}
\begin{equation}
\begin{split}
\vec{c_{\backslash t}}=&Relu(W_2\cdot Relu(W_1\cdot[\vec{h_l}:\cev{h_r}]+b_1)+b_2)
\end{split}
\end{equation}
\end{small}
Let $SW$ be all synsets occurred in the training corpora. Let $s_t$ be the correct synset for $w_t$, the probability distribution $p(s_t|{c \backslash \{w_t\}})$ over all the synsets $s_i \in SW$ is calculated by a softmax layer:
\begin{small}
\begin{equation}
p(s_t|c \backslash \{w_t\})=\frac{exp(\vec{s_t}^ {\mathrm{ T }}\cdot\vec{c_{\backslash t}})}{\sum\limits_{s_i\in SW}exp(\vec{s_i}^ {\mathrm{ T }}\cdot\vec{c_{\backslash t}})}
\label{p1}
\end{equation}
\end{small}
For a word $w$ in context $c$, the probability distribution $p(w|c \backslash \{w\})$ over all words $w'$ is calculated by another softmax layer:
\begin{small}
\begin{equation}
p(w|c \backslash \{w\})=\frac{exp(\vec{w}^ {\mathrm{ T }}\cdot\vec{c_{\backslash t}})}{\sum\limits_{w'}exp(\vec{w'}^ {\mathrm{ T }}\cdot\vec{c_{\backslash t}})}
\label{p2}
\end{equation}
\end{small}
In the labeled corpus $T=\{c_1, c_2, ...,c_N\}$, for each sentence $c_j\in T$, the training objective is:
\begin{small}
\begin{equation}
    J_a=-\sum_{c_j}(log(p(s_t|c_j \backslash \{w_t\})))
\label{ja}
\end{equation}
\end{small}
The training process for \eqref{ja} is illustrated in Figure \ref{fig1}.

As an unsupervised task, in the unlabeled corpus, the model is trained to make prediction of word $w$ from context $c_j\backslash\{w\}$.  The objective is illustrated as Figure \ref{fig2} and formulated by:
\begin{small}
\begin{equation}
\label{ju}
\begin{split}
J_u=-\sum_{c_j}\sum_{ w \in c_j \cap FW} (log(p(w|c_j\backslash \{w\})))
\end{split}
\end{equation}
\end{small}
where $\theta$ is an adjustable parameter. The final objective $J$ is the weighted aggregation of $J_a$ and $J_u$:
\begin{small}
\begin{equation}
J=J_a+\alpha J_u
\label{j}
\end{equation}
\end{small}

where $\alpha$ is an adjustable parameter. $J$ is trained to be minimized and equation \eqref{p1} and \eqref{p2} are approximated by sampled softmax \citep{DBLP:conf/acl/JeanCMB15} during training.
 During inferencing, for $w_t$ in $c_j$ to be disambiguated, the model chooses the candidate synset with  highest probability conditioned on $c_j$ as output ($s_o$), which is formulated by:
\begin{small}
\begin{equation}
s_o=\arg\max_{s_t}(p(s_t|c_j\backslash \{w_t\})),s_t \in S(w_t)
\end{equation}
\end{small}
where $S(w_t)$ is the set of candidate synsets for $w_t$, $p(s_t|c_j\backslash \{w_t\})$ is the  outputs of our BLSTM encoder.

\section{Experiments}

\subsection{Setup}
\textbf{Baseline Methods:} The baselines include some state-of-the-art approaches, i.e., MFS (to directly output the Most Frequent Sense in WordNet); IMS \citep{Zhi2010It}, a classifier working on several hand-crafted features, i.e., POS, surrounding words and local collocations; Babelfy \citep{Moro2014Entity}, a state-of-the-art knowledge-based WSD system exploiting random walks to connect synsets and text fragments; Lesk\_ext+emb \citep{Basile2014An}, an extension of Lesk by incorporating similarity information of definitions; UKB\_gloss \citep{Agirre2009Personalizing,DBLP:journals/coling/AgirreLS14}, another graph-based method for WSD; A joint learning model for WSD and entity linking (EL) utilizing semantic resources by \citet{DBLP:conf/acl/WeissenbornHXU15};  IMS{-s}+emb \citep{Iacobacci2016Embeddings}, the combination of original IMS and word embeddings through exponential decay while surrounding words are removed from features; Context2vec \citep{DBLP:conf/conll/MelamudGD16}, a generic model for generating representation of context for WSD; Jointly training LSTM with labeled and unlabeled data \citep{DBLP:journals/corr/abs-1712-03376} (this is an open implementation for part of the work of \citet{DBLP:conf/coling/YuanRDEA16}. Since the models and 100 billion data used in Yuan et al.'s paper are not available, we select Le et al.'s work as an alternative. Le et al.'s work uses 1 billion unlabeled data, which is roughly equal to the size of unlabeled corpus in our work. This makes the comparison more fair); A model jointly learns to predict word senses, POS and coarse-grained semantic labels by \citet{Raganato2017Neural}; Train-O-Matic \citep{pasini-navigli:2017:EMNLP2017}, a language-independent approach for generating sense-labeled data automatically based on random walk in WordNet and training a classifier on it.
\begin{table}
    \centering
    \begin{tabular}{|c ||c | } \hline
                  Methods&  Concatenation of All Test Sets\\ \hline
            MFS &  65.8  \\ \hline
            Babelfy & 66.4\\ \hline
            UKB\_gloss & 61.1\\ \hline
            Lesk\_ext+emb & 64.2\\ \hline
            \multicolumn{2}{|c|}{Ours}\\ \hline
            DisDict& 66.2\\ \hline
            DisDict+MFS & \textbf{67.2}\\ \hline
    \end{tabular}
    \caption{F1-scores (\%) for  English all-words fine-grained WSD in the absence \\of manually labeled training data}
    \label{result_backoff}
\end{table}

\begin{table}
    \centering
    \begin{tabular}{|c||c| c | c |c|} \hline
                  \multirow{2}*{Methods}&  \multicolumn{4}{|c|}{Test Datasets} \\ \cline{2-5}
            & SE2 &  SE3 & SE13 & SE15  \\ \hline
            MFS & 72.0 & 72.0 & 63.0 & 66.3  \\ \hline
            OMSTI & 73.3& 67.5 & 62.5 & 63.4\\ \hline
            Train-O-Matic & 71.1 & 67.8 & 65.8& 68.1   \\ \hline
            \citet{DBLP:conf/acl/WeissenbornHXU15} & - & 68.8 & \textbf{72.8} & 71.5  \\ \hline
            \multicolumn{5}{|c|}{Ours}\\ \hline
            DisDict & 73.2  & 69.0 & 66.2 &  70.1 \\ \hline
            DisDict+MFS & 74.6 & 72.0 & 65.3 & 71.0  \\ \hline
            MLab+DisDict+ULab+MFS & \textbf{78.0} & \textbf{76.0} & 70.9 & \textbf{75.1}  \\ \hline
    \end{tabular}
    \caption{F1-scores (\%) for nouns disambiguation}
    \label{result_noun}
\end{table}
\textbf{Datasets:} We choose \textit{Semcor 3.0} \citep{DBLP:conf/naacl/MillerCLLT94} (226,036 manual sense annotations), which is also used by baselines, as the manually labeled data. We also extract 27,616,880 word-context pairs from Wikipedia April 2010 dump with 1 billion tokens which was preprocessed and utilized by \cite{Sun2016Inside}. From which, we generate 11,925,166 sense labeled instances by DisDict. The trained models are evaluated on the fine-grained English all-words WSD task under the standardized evaluation framework released by \citet{DBLP:conf/eacl/NavigliCR17}. We tune parameters on \textit{SemEval-07 task 17} \citep{pradhan-EtAl:2007:SemEval-2007} and test models on four datasets, i.e., \textit{senseval-2} \citep{edmonds-cotton:2001:SENSEVAL} with 2282 synset annotations, \textit{senseval-3 task 1} \citep{snyder-palmer:2004:Senseval-3} with 1850 annotations, \textit{SemEval-13 task 12} \citep{navigli-jurgens-vannella:2013:SemEval-2013} with 1644 annotations and \textit{SemEval-15 task 13} \citep{moro-navigli:2015:SemEval} with 1022 annotations. The word embeddings we use are pretrained on 2 billion ukWac \citep{DBLP:journals/lre/BaroniBFZ09} corpus, the same corpus as that used in baseline methods.
\begin{figure*}[t]
\centering
\includegraphics[width = 150 mm, height = 70 mm]{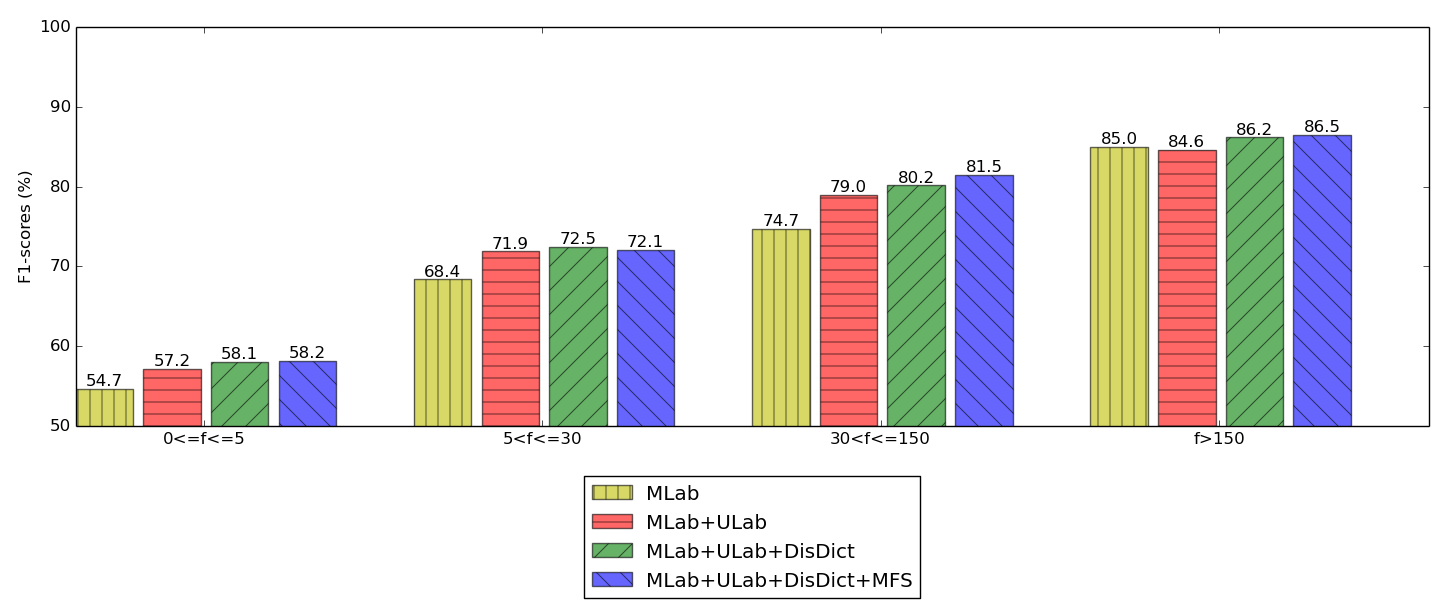}
\caption{F1-scores (\%) for test synsets grouped by their frequencies (f) in the manually \\labeled training corpus}\label{freq}
\end{figure*}

\textbf{Settings:} We design a series of experiments based on the combination of four basic settings, i.e., \textit{MLab} (conducting supervised learning on manually labeled data); \textit{ULab} (conducting unsupervised on unlabeled data); \textit{DisDict} (conducting supervised  learning on data generated by DisDict); \textit{MFS} (adding a bias towards the output score of most frequent synset when inferencing). For \textit{MLab}, if with \textit{MFS}, we select it as the backoff strategy when the target word is unseen in the training corpora; elsewise, we randomly select a candidate synset to output under such circumstance. For the combination of \textit{MLab} and \textit{DisDict}, during training, we sample instances from the two datasets  with a ratio to control the balance.

For data generation, $N_f$ is set as $10$, $\tau$ in \eqref{pmi} is set as 0.66 and $p$ is set as 0.3 in \eqref{freq}; For training, the BLSTM has 2 layers and 400 hidden units for each individual LSTM; the learning rate for $J$ in \eqref{j} is set as $0.1$ and $\alpha$ is $5.0$, we use Mini-batch Gradient Descent to optimize $J$, and the batch size is 80; we apply Dropout \citep{DBLP:journals/jmlr/SrivastavaHKSS14} with a rate of 0.5 to prevent over-fitting; the dimension of word and synset embedding is 400; the ratio of samples from manually labeled data and that generated by DisDict is set as $1.0:0.3$; During inferencing, for $MFS$, the bias added to the most frequent synset is set as 0.5.

\subsection{Results and Analysis}
Table \ref{result} posts the results for English all-words fine-grained WSD with the utilization of manually labeled training data. From the comparison between  \textit{MLab} and \textit{MLab+ULab} setting, it shows that unsupervised learning improves the generalization performance. From the comparison between  \textit{MLab+ULab} and \textit{MLab+DisDict+ULab} setting, it illustrates the knowledge in DisDict has a good quality and the combination of knowledge and unlabeled data really generates reliable sense-labeled data, thus improves the overall performance. From the comparison between  \textit{MLab+DisDict+ULab} and \textit{MLab+DisDict+ULab+MFS} setting, it shows that \textit{MFS} has a good complementarity to supervised WSD.

Table \ref{result_backoff} reports the results of WSD in the absence of manually labeled training data. For fairness, in Table \ref{result_backoff}, we only make comparison with methods do not need manually labeled data. As shown in Table \ref{result_backoff}, when there is no manually-labeled data, the combination of our model and \textit{MFS}, i.e., \textit{DisDict+MFS}, has outperformed the \textit{MFS} and \textit{Babelfy} baseline, thus provides a potential solution for similar tasks in a  lack of manual annotations.

Table \ref{result_noun} reports the results of nouns disambiguation on several test sets. Among these settings, \textit{MFS}, \textit{OMSTI}, \textit{Train-O-Matic}, \textit{DisDict} and \textit{DisDict+MFS} do not require manually labeled data while \citet{DBLP:conf/acl/WeissenbornHXU15} and \textit{MLab+DisDict+ULab+MFS} require. For data generation methods which are closet to ours, i.e., \textit{OMSTI} and \textit{Train-O-Matic}, we train our supervised model on their generated data and post the results.  Compared them with \textit{DisDict}, we can concludes that our new data generation method beats these two approaches by two points. First, these methods only focus on nouns disambiguation, while our work is appropriate to any POS in English. Second, our generated data has a higher quality with a better overall performance on nouns disambiguation.

For fine-grained analysis, we divide synsets in the test sets into four groups according to their frequencies $f$ in the manually labeled training corpus, i.e., $0<=f<=5$, $5<f<=30$, $30<f<=150$, $f>150$, and calculate F1-scores for these groups, as shown in Figure \ref{freq}. It shows that incorporating data generated by DisDict is beneficial consistently to synsets in each frequency interval from the comparison between \textit{MLab+ULab} and \textit{MLab+DisDict+ULab}.

\section{Conclusion and Future Work}

In this paper, we propose a new framework
to combine supervised learning and knowledge-based approaches for word sense disambiguation
(WSD). Under this framework, we automatically construct a semantic KB, i.e.,
DisDict, to highlight the semantic differences among synsets, and utilize DisDict to generate reliable sense-labeled data from unlabeled corpora. Then we apply a neural model to conduct  both supervised and unsupervised learning for WSD. Evident from the experiments, our framework outperforms a
number of representative approaches on major standard evaluation datasets. Furthermore, our model also achieves better performance against other methods when there is no manually labeled data, thus provides a potential solution for such learning tasks in a lack of manually annotations.

For future work, we will focus on two research lines. First, we will study
more powerful approaches to acquire knowledge from unstructured
data automatically. Second, we will study a better combination of the data manually labeled and that generated by DisDict.
\newcommand{\tabincell}[2]{\begin{tabular}{@{}#1@{}}#2\end{tabular}}

\bibliographystyle{aaai}\bibliography{zhong,delli,xiong,neale,socher,nee,nieto,bovi,banerjee,basile,miller,agirre,moro,lesk,ims,bilstm,seq,plus_emb,semi,concept5,retrofit,wordnet,keok,context2vec,att_rel_pred,grammar,attention,seq2seq,LSTM,bilstm2,gen_seq,semantic_role,seq_label,parsering,ext_lesk,unified_rep,multiling1,game,randomwalk,guo,multi_obj,semcor,OMITS,corpus,senseval2,senseval3t1,senseval07t17,semeval13t12,semeval15t13,conceptnetWSD,babelnet,feature_selection,ukwac,trainomatic,sampledsoftmax,conceptnet,100b,dropout,sun,knowledge2,auto_extend,causalnet,pmi,fastsense}

\end{document}